\documentclass[11pt, a4paper, oneside]{Thesis} 
\usepackage{multirow}
\graphicspath{{Pictures/}} 

\usepackage[backend=bibtex]{biblatex}
\title{\ttitle} 

\begin{document}

\frontmatter 

\setstretch{1.3} 

\fancyhead{} 
\rhead{\thepage} 
\lhead{} 

\pagestyle{fancy} 

%

\thesistitle{Improving Surrogate Gradient Learning in Spiking Neural Networks via Regularization and Normalization}
\documenttype{Undergraduate Thesis}
\supervisor{Dr. Timothée Masquelier}
\supervisorposition{Senior Research Scientist}
\supervisorinstitute{CERCO, CNRS, France}
\cosupervisor{Dr. FirstName \textsc{SecondName}}
\cosupervisorposition{Asst. Professor}
\cosupervisorinstitute{BITS-Pilani Hyderabad Campus}
\examiner{}
\degree{Bachelor of Engineering in Computer Science, Master of Science in Mathematics}
\coursecode{BITS F421T}
\coursename{Thesis}
\authors{M S Nandan}
\IDNumber{2017B4A70657G}
\addresses{}
\subject{}
\university{\texorpdfstring{\href{http://www.bits-pilani.ac.in/} 
                {Birla Institute of Technology and Science Pilani, Goa Campus}} 
                {Birla Institute of Technology and Science Pilani, Goa Campus}}
\UNIVERSITY{\texorpdfstring{\href{http://www.bits-pilani.ac.in/} 
                {BIRLA INSTITUTE OF TECHNOLOGY AND SCIENCE PILANI, GOA CAMPUS}} 
                {BIRLA INSTITUTE OF TECHNOLOGY AND SCIENCE PILANI, GOA CAMPUS}}
\department{\texorpdfstring{\href{http://www.bits-pilani.ac.in/pilani/computerscience/ComputerScience} 
                {Computer Science \& Information Systems}} 
                {Computer Science}}
\DEPARTMENT{\texorpdfstring{\href{http://www.bits-pilani.ac.in/pilani/computerscience/ComputerScience} 
                {COMPUTER SCIENCE \& INFORMATION SYSTEMS}} 
                {COMPUTER SCIENCE \& INFORMATION SYSTEMS}}
\group{\texorpdfstring{\href{Research Group Web Site URL Here (include http://)}
                {Research Group Name}} 
                {Research Group Name}}
\GROUP{\texorpdfstring{\href{Research Group Web Site URL Here (include http://)}
                {RESEARCH GROUP NAME (IN BLOCK CAPITALS)}}
                {RESEARCH GROUP NAME (IN BLOCK CAPITALS)}}
\faculty{\texorpdfstring{\href{Faculty Web Site URL Here (include http://)}
                {Faculty Name}}
                {Faculty Name}}
\FACULTY{\texorpdfstring{\href{Faculty Web Site URL Here (include http://)}
                {FACULTY NAME (IN BLOCK CAPITALS)}}
                {FACULTY NAME (IN BLOCK CAPITALS)}}

\maketitle

\begin{abstract}
Spiking neural networks (SNNs) are different from the classical networks used in deep learning: the neurons communicate using electrical impulses called spikes, just like biological neurons. SNNs are appealing for AI technology, because they could be implemented on low power neuromorphic chips. However, SNNs generally remain less accurate than their analog counterparts. In this report, we examine various regularization and normalization techniques with the goal of improving surrogate gradient learning in SNNs. 
\end{abstract}

\begin{acknowledgements}
I would like to express my sincere gratitude for the support of my supervisor, Dr. Timothée Masquelier, and for granting me the opportunity to pursue my undergraduate thesis under his guidance. I would also like to acknowledge my university, BITS Pilani, Goa, for their support and help during my thesis.
\end{acknowledgements}


\pagestyle{fancy}

\lhead{\emph{Contents}} 
\tableofcontents 

\lhead{\emph{List of Figures}}
\listoffigures 

\lhead{\emph{List of Tables}}
\listoftables 

\mainmatter 

\pagestyle{fancy} 



\chapter{Introduction} 

\label{Chapter1} 

\lhead{Chapter 1. \emph{Introduction}} 

Spiking neural networks (SNNs) are a type of neural network that mimic natural neural networks more closely than traditional artificial neural networks (ANNs). SNNs operate using spikes, which are discrete events that take place at points in time, rather than continuous values.
The occurrence of a spike is determined by differential equations that represent various biological processes. Essentially, once the membrane potential of a neuron reaches a certain threshold, it emits a spike, and the potential of that neuron is reset. These spikes are sent as signals to other neurons which, in turn, increase or decrease their potentials in response to these signals.

SNNs, which could be implemented on low power neuromorphic chips such as Intel Loihi \cite{davies2018loihi}, are regarded as a potential competitor of ANNs due to their high biological plausibility, event-driven property, and low power consumption. However, SNNs generally remain less accurate than ANNs. In recent years, SNNs have attracted the deep learning community since the breakthrough of surrogate gradient learning \cite{neftci2019surrogate}, which enabled the training of networks with backpropagation despite the non-differentiable condition for spike emission.

Like their analog counterpart, SNNs consist of a large number of parameters. A higher number of parameters gives neural networks the power to fit multiple types of datasets. However, this can lead to a model learning the noise in the training data to the extent that it negatively impacts the performance of the model on new data. This is known as overfitting, and it is one of the most common issues faced while training a model.

In this report, we examine different regularization techniques, namely weight decay and spike penalization, to see if they help reduce overfitting in SNNs. We also examine the benefits of weight normalization when used to train SNNs. We use the CIFAR10 data-set \cite{krizhevsky2009learning} in all our experiments. This is a relatively small data-set with high complexity and models trained on it tend to overfit, which makes it suitable for our experiments.


\chapter{Weight Decay} 

\label{Chapter2} 

\lhead{Chapter 2. \emph{Weight Decay}} 

Weight decay is a regularization technique in which a small penalty, the $L_2$ norm of the weights of the model, are added to the loss function.
\begin{equation}
    L_{new}(w) = L_{old}(w) + \lambda \|w\|^2
\end{equation}

If a model has overfit the training data, it means that the model fits exactly against its training data. This could happen when the model trains for too long on the training data or when the model is too complex, in which case the model memorizes the data and fits too closely to the training set. Such a model may give a small error on the training set but it cannot perform accurately against unseen data. Weight decay can help prevent this as the additional penalty term controls the excessively fluctuating function such that the coefficients do not take extreme values.

\section{Experiments}

Weight decay is a popular technique used to prevent overfitting in ANNs \cite{xie2020understanding}. However, its benefits when used with SNNs has not been investigated thoroughly. In this report, we examine the benefits of weight decay in SNNs by testing it with different architectures.


\subsection{Spiking Convolutional Neural Network}
\label{section:2.1}

The first type of SNN on which we tried weight decay was a simple spiking convolutional neural network. The configuration of this model is $\{c128k3s1\text{-}BN\text{-}IF\text{-}MPk2s2\}\text{*}4\text{-}FC$. Here $c128k3s1$ denotes a convolutional layer with number of channels = 128, kernel size = 3 and stride = 1, $BN$ denotes a batch normalization layer, $IF$ denotes Integrate-and-Fire neurons, $MPk2s2$ denotes a max pooling with kernel size 2 and stride 2. The symbol $\{\}\text{*}4$ denotes 4 repeated structures, and $FC$ denotes a fully connected layer. The model was trained for 100 epochs using the SGD (stochastic gradient descent) optimizer with a learning rate of 0.1 and a momentum of 0.9. The CosineAnnealingLR scheduler was also used with $T_{max}$ = total number of epochs. The results are shown in Table \ref{table:1}.
\\
\begin{table}[h!]
\centering
 \begin{tabular}{|c|c|c|} 
 \hline
 Weight decay & Training Accuracy (\%) & Testing Accuracy (\%)\\ 
 \hline
 0 & 94.24 & 86.95 \\ 
 0.0001 & 95.34 & 88.13 \\
 0.0003 & 94.42 & \textbf{88.53} \\
 0.0005 & 93.35 & 88.19 \\
 \hline
 \end{tabular}
\caption{Comparison of spiking convolutional neural network models trained with SGD and different weight decay coefficients.}
\label{table:1}
\end{table}

We performed a similar experiment using the AdamW \cite{loshchilov2017decoupled} optimizer with a learning rate of 0.01. The results are shown in Table \ref{table:2}.
\\
\begin{table}[h!]
\centering
 \begin{tabular}{|c|c|c|} 
 \hline
 Weight decay & Training Accuracy (\%) & Testing Accuracy (\%)\\ 
 \hline
 0 & 95.06 & 86.91 \\ 
 0.0003 & 95.40 & 87.07 \\
 0.003 & 95.67 & 87.47 \\
 0.03 & 93.82 & \textbf{88.33} \\
 0.3 & 74.15 & 74.41 \\
 \hline
 \end{tabular}
\caption{Comparison of spiking convolutional neural network models trained with AdamW and different weight decay coefficients.}
\label{table:2}
\end{table}

From Table \ref{table:1} and Table \ref{table:2}, it can be seen that with the optimal coefficient, weight decay can reduce overfitting and help improve testing accuracy when used with both SGD and AdamW. 

\subsection{Spike-Element-Wise (SEW) ResNet}

In neural networks, multiple layers are used to learn representations of data with multiple levels of abstraction. Deeper networks have advantages over shallower networks in terms of computation cost and generalization ability. However, deep networks are hard to train because of the vanishing gradient problem: as the gradient is backpropagated to earlier layers, repeated multiplication may make the gradient vanishingly small. As a result, as the network goes deeper, its performance gets saturated or even starts degrading rapidly. To solve this problem, residual blocks were proposed \cite{he2016deep}. In these blocks, skip connections are used, which skip training from a few layers and connect directly to the output.

The Spike-Element-Wise (SEW) ResNet was introduced in \cite{fang2021deep} to realize residual learning in SNNs. In this report, we examine the benefits of weight decay in SEW ResNets. For our experiments, we used the architecture $Conv\text{-}BN\text{-}IF\text{-}\{SEW Block\text{-}SEW Block\text{-}MPk2s2\}\text{*}5\text{-}FC10$, where the $SEW Block$ consists of two convolutional layers with ADD as the element-wise function $g$. For the convolutional layers, we experimented with both 32 and 64 channels. We used SGD with learning rates 0.1 and 0.001, and a momentum of 0.9. The CosineAnnealingLR scheduler was also used with $T_{max}$ = total number of epochs.
\\
\begin{table}[h!]
\centering
 \begin{tabular}{|c|c|c|c|c|} 
 \hline
 Weight & \multicolumn{2}{c|}{Channels=32} & \multicolumn{2}{c|}{Channels=64} \\
 \cline{2-5}
 decay & Train Accuracy (\%) & Test Accuracy (\%) & Train Accuracy (\%) & Test Accuracy (\%)\\
 \hline
 0 & 61.11 & 61.12 & 54.19 & 55.05 \\ 
 0.00003 & 78.73 & 78.38 & 80.17 & 78.77 \\
 0.0003 & 82.99 & \textbf{81.47} & 88.69 & \textbf{85.34} \\
 0.003 & 65.07 & 65.81 & 72.49 & 71.52 \\
 \hline
 \end{tabular}
\caption{Comparison of SEW ResNet models trained with SGD, learning rate=0.1 and different weight decay coefficients.}
\label{table:3}
\end{table}
\\
\begin{table}[h!]
\centering
 \begin{tabular}{|c|c|c|c|c|} 
 \hline
 Weight & \multicolumn{2}{c|}{Channels=32} & \multicolumn{2}{c|}{Channels=64} \\
 \cline{2-5}
 decay & Train Accuracy (\%) & Test Accuracy (\%) & Train Accuracy (\%) & Test Accuracy (\%)\\
 \hline
 0 & 85.85 & 83.26 & 93.28 & 85.94 \\ 
 0.00003 & 86.07 & 82.78 & 93.19 & \textbf{87.58} \\
 0.0003 & 86.46 & \textbf{83.58} & 93.39 & 87.17 \\
 0.003 & 86.30 & 83.53 & 93.30 & 87.19 \\
 \hline
 \end{tabular}
\caption{Comparison of SEW ResNet models trained with SGD, learning rate=0.001 and different weight decay coefficients.}
\label{table:4}
\end{table}

From Tables \ref{table:3} and \ref{table:4}, we can see that weight decay helps in testing improving accuracy and hence, reducing overfitting. In Table \ref{table:3}, the large increase in accuracy on using weight decay shows that weight decay can sometimes be important for the model to learn.   

\subsection{Spiking ConvMixer}

The ConvMixer architecture, introduced in \cite{anonymous2022patches}, consists of a patch embedding layer followed by repeated applications of a simple fully-convolutional block. These blocks consist of depthwise convolution (i.e., grouped convolution with groups equal to the number of channels) followed by pointwise (i.e., kernel size 1×1) convolution. After this, global pooling is performed, followed by a fully connected layer.

In this report, we implemented a spiking version of the ConvMixer architecture by replacing the GELU activation function with IF neurons and by replacing the residual block with a Spike-Element-Wise (SEW) residual block. The optimal hyperparameters used for our experiments were width (number of channels in the convolutional layers) = 256, depth (number of repetitions of the ConvMixer layer) = 8, patch size = 1 and kernel size = 9. We then tested weight decay on this architecture. We used SGD with learning rate of 0.1 and a momentum of 0.9. The CosineAnnealingLR scheduler was also used with $T_{max}$ = total number of epochs. The results are shown in Table \ref{table:8}.
\\
\begin{table}[h!]
\centering
 \begin{tabular}{|c|c|c|} 
 \hline
 Weight decay & Training Accuracy (\%) & Testing Accuracy (\%)\\ 
 \hline
 0 & 99.79 & 91.00 \\ 
 0.00001 & 99.79 & 91.37 \\
 0.0001 & 99.54 & \textbf{92.53} \\
 0.0005 & 94.74 & 89.28 \\
 \hline
 \end{tabular}
\caption{Comparison of Spiking ConvMixer models trained with SGD and different weight decay coefficients.}
\label{table:8}
\end{table}

From Table \ref{table:8}, we can see that weight decay can reduce overfitting in Spiking ConvMixer models and help improve accuracy.

\chapter{Spike Penalization} 

\label{Chapter3} 

\lhead{Chapter 3. \emph{Spike Penalization}} 

Energy efficiency is a desirable property of neural network models. For SNNs, this would mean that their neuron spiking activity should be as sparse as possible, while still performing the task with high accuracy. This property is also desirable from a biological point of view, since biological neurons are very energy efficient and emit limited amounts of spikes in a given amounts of time. If sparse, patterns of activity might also be more explainable. 

In order to enforce sparse spiking activity, we add the following term to the loss function:
\begin{equation} \label{eq:1}
    L_r(l) = \frac{1}{2KN} \sum_n \sum_k S_k^2 [n]
\end{equation}
where $l$ is the layer, $K$ is the number of neurons and $N$ is the number of time steps. $S_k[n]$ denotes whether the $k^{th}$ neuron spiked at the $n^{th}$ time-step, i.e., $S_k[n] = 1$ if the $k^{th}$ neuron spiked at the $n^{th}$ time-step, else $S_k[n] = 0$. $S_k^2 [n]$ is used instead of $S_k[n]$ in order to ensure that the regularization will not be applied to neurons that have not emitted any spikes, as explained in \cite{pellegrini2021low}. Sparse activity may also lead to regularization and hence, lead to less overfitting. In this report, we test this premise. 

\section{Experiments}

For the experiments, we use the same model that we had used in Section \ref{section:2.1}. The models were trained with SGD with a learning rate of 0.1 and a momentum of 0.9. The CosineAnnealingLR scheduler was also used with $T_{max}$= total number of epochs. We then train the models by adding the term mentioned in Equation \ref{eq:1} multiplied by the spike penalization coefficient to the loss function. The results are shown in Table \ref{table:7}.


\begin{table}[h!]
\centering
 \begin{tabular}{|c|c|c|c|c|} 
 \hline
 Spike penalization & \multicolumn{2}{c|}{Training} & \multicolumn{2}{c|}{Testing} \\
 \cline{2-5}
 weight & Spike Rate & Accuracy & Spike Rate & Accuracy \\
 \hline
    0 & 0.1118 & 94.47 & 0.1003 & 87.04 \\
    0.01 & 0.1074 & 94.58 & 0.0951 & 86.82 \\
    0.05 & 0.0937 & 94.57 & 0.0830 & \textbf{87.32} \\
    0.1 & 0.0767 & 94.56 & 0.0676 & 87.22 \\
    0.5 & 0.0429 & 94.12 & 0.0378 & 86.93 \\
    1 & 0.0275 & 93.34 & 0.0244 & 86.6 \\
    1.5 & 0.0208 & 92.53 & 0.0183 & 86.13 \\
    2 & 0.0178 & 91.87 & 0.0157 & 85.66 \\
    2.5 & 0.0154 & 91.38 & 0.0136 & 85.6 \\
    5 & 0.0093 & 88.07 & 0.0083 & 83.75 \\
    10 & 0.0059 & 84.27 & 0.0053 & 82.11 \\
 \hline
 \end{tabular}
\caption{Comparison of spiking convolutional neural network models trained with different spike penalization weights.}
\label{table:7}
\end{table}

From Table \ref{table:7}, we can see that the spike rate decreases with an increase in the spike penalization weight. However, the best accuracy is obtained when the spike penalization weight is 0.05, which shows that penalizing the spikes can help reduce overfitting. 

\begin{figure}[h]
\includegraphics[width=12cm]{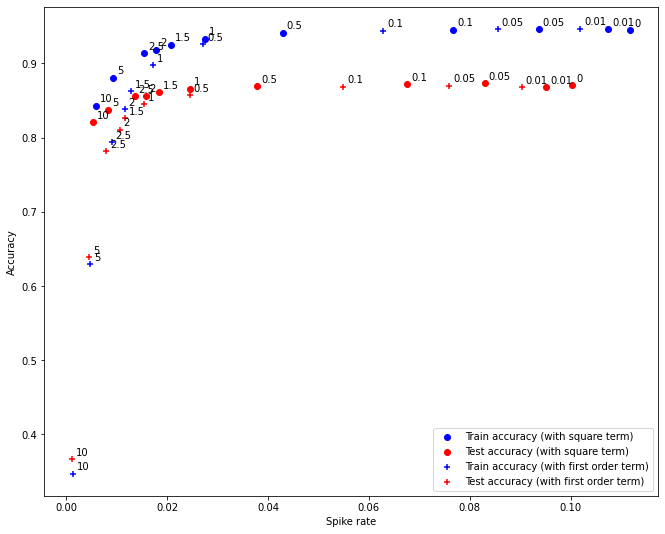}
\centering
\caption{Plot of training and testing accuracies vs spike rate of spiking convolutional neural network models trained with different spike penalization loss terms.}
\label{fig:1}
\end{figure}

We also perform experiments to analyze whether using a square term ($S_k^2 [n]$) in the loss term in Equation \ref{eq:1} gives better results than using a first order term ($S_k[n]$). The plot of the testing accuracies are shown in Figure \ref{fig:1}. 

From Figure \ref{fig:1}, we can see that when the spike penalization weights are small, both loss terms give similar results. However, when the weights are larger, the drop in accuracy is too large when compared to the decrease in spike rate for the loss term with the first order term. Hence, the loss term with the square term is better.  

\chapter{Weight Normalization} 

\label{Chapter4} 

\lhead{Chapter 4. \emph{Weight Normalization}} 

Weight Normalization was introduced in \cite{salimans2016weight} in order to help speed up convergence of stochastic gradient descent. This method was introduced as an alternative to batch normalization. The authors of the paper claim that although the method is simpler, it provides much of the speed-up of full batch normalization with lower computational overhead. 

In this method, each weight vector \textbf{w} of the neural network is reparameterized in terms of a parameter vector \textbf{v} and a scalar parameter $g$. This is done by expressing \textbf{w} in terms of the new parameters as follows:
\begin{equation}
    \textbf{w} = \frac{g}{||\textbf{v}||}\textbf{v}
\end{equation}
Stochastic gradient descent is performed with respect to these new parameters instead of the original weight vector. The authors claim that decoupling the norm of the weight vector ($g$) from the direction of the weight vector (\textbf{v}/$||\textbf{v}||$) improves the conditioning of the gradient and leads to improved convergence of the optimization procedure.

Since batch normalization has the benefit of fixing the scale of the features generated by each layer of the neural network and since weight normalization lacks this property, the authors believe that it is important to properly initialize our parameters, and have introduced a data-dependent method for doing so. The authors also explore the idea of combining weight normalization with a special version of batch normalization, called mean-only batch normalization, where the minibatch means are subtracted like with full batch normalization, but division by the minibatch standard deviations is not performed.  Mean-only batch normalization has the effect of centering the gradients that are backpropagated. The computational overhead of mean-only batch normalization is lower than that of full batch normalization. 

\section{Experiments}

In this report, we explore the benefits of weight normalization, as well as weight normalization + mean-only batch normalization (with affine transform) when used while training SNNs. For the experiments, we use the same model that we had used in Section \ref{section:2.1}. The models were trained with SGD with a learning rate of 0.1 and a momentum of 0.9. The CosineAnnealingLR scheduler was also used with $T_{max}$= total number of epochs. For each case, we pick the optimal weight decay coefficient from {0, 0.0001, 0.0003}. The results are shown in Table \ref{table:5}.
\\
\begin{table}[h!]
\centering
 \begin{tabular}{|c|c|c|} 
 \hline
 Normalization method & Training Accuracy (\%) & Testing Accuracy (\%)\\
 \hline
 Batch normalization & 94.42 & 88.53 \\ 
 Weight Normalization & 93.17 & 87.83 \\
 Weight Normalization + mean-only & 95.55 & \textbf{88.54} \\
  batch normalization & & \\
 \hline
 \end{tabular}
\caption{Comparison of spiking convolutional neural network models trained with different normalization methods.}
\label{table:5}
\end{table}

From Table \ref{table:5}, we can see that weight normalization + mean-only batch normalization can help improve the accuracy of the network. However, the improvement when compared to the model trained with batch normalization is not significant.

We also evaluate the benefits of the data-dependent initialization method introduced in the paper. We use the same model as used in the experiment above. The model is trained using weight normalization. The optimizer is SGD with a learning rate of 0.1, momentum of 0.9 and no weight decay. The CosineAnnealingLR scheduler was also used with $T_{max}$= total number of epochs. The results are shown in Table \ref{table:6}. 
\\
\begin{table}[h!]
\centering
 \begin{tabular}{|c|c|c|} 
 \hline
 Model & Training Accuracy (\%) & Testing Accuracy (\%)\\
 \hline
 With data-dependent initialization & 86.20 & 81.98 \\ 
 Without data-dependent initialization & 91.09 & \textbf{86.15} \\
 \hline
 \end{tabular}
\caption{Comparison of spiking convolutional neural network models trained with and without data-dependent initialization.}
\label{table:6}
\end{table}

From Table \ref{table:6}, we can see that the data-dependent initialization method does not work well for our model. Hence, we can conclude that the data-dependent initialization method, which works for ANNs, might not be suitable for SNNs. 

\chapter{Conclusion} 

\label{Chapter5} 

\lhead{Chapter 5. \emph{Conclusion}} 

Spiking neural networks have risen in popularity over the past few years, especially since the introduction of surrogate gradient learning \cite{neftci2019surrogate}. However, the performance of SNNs still lags behind that of ANNs. In this report, we have examined various techniques, namely weight decay, spike penalization and weight normalization, in order to improve the performance of SNNs. Our experiments show that the first two of these techniques can indeed improve the accuracy of SNNs. 








\backmatter


\label{Bibliography}

\lhead{\emph{Bibliography}} 

\printbibliography

\end{document}